\documentclass{llncs}

\usepackage{amsmath}

\usepackage{graphics}
\usepackage{verbatim}
\usepackage{epsfig}
\usepackage{eso-pic,graphicx}
\usepackage{setspace}
\usepackage{subfigure}
\usepackage{fancyhdr}

\usepackage{makeidx}
\usepackage{multirow}
\usepackage{color}

\begin{document}

%

 \pagestyle{fancy}
\lhead{ Scale Space and Variational Methods 2011 (to appear)} 
\rhead{}

\title{From a Modified Ambrosio-Tortorelli to a Randomized Part Hierarchy Tree}

\author{ Sibel Tari \and Murat Genctav}
\institute{Middle East Technical University, Department of
Computer Engineering, \\ Ankara, TR-06531\\
\email{stari@metu.edu.tr\footnote{Corresponding author}}  \email{muratgenctav@gmail.com} }

\maketitle

\thispagestyle{fancy}

\begin{abstract}
We demonstrate the possibility of coding parts, features that are higher level than boundaries, using a modified AT field after augmenting the interaction term of the AT energy with a non-local term and weakening the separation into boundary/not-boundary phases. The  iteratively extracted parts  using the level curves with double point singularities are organized as a proper binary tree. Inconsistencies due to non-generic configurations for level curves   as well as  due to visual changes such as occlusion  are successfully handled once  the tree is endowed with a probabilistic structure. The work is a step in establishing  the AT function as a bridge between low and high level visual processing. \\

\end{abstract}

\section{Introduction}

The phase field  of Ambrosio and Tortorelli~\cite{Ambrosio90} (AT function) serving as a  continuous indicator for the boundary/not-boundary state at every domain point has proven to be an indispensable tool in image and shape analysis. 
It is a minimizer of an energy composed of two competing terms: One term  favors  configurations that take values close to either $0$ or $1$ (separation into boundary/not-boundary phases) and the other term  encourages local interaction in the domain by penalizing spatial inhomogeneity.     A  parameter controls the relative influence of these two terms, hence, the interaction.   As this "interaction"  parameter tends to $0$, the  separation term  is strongly emphasized; consequently, the field  tends  to the characteristic function $1-\chi_S$ of the boundary set $S$ and the AT energy tends  (following the $\Gamma$ convergence framework~\cite{Braides98})  to the boundary length. 

 In computer vision, the AT  function  first appeared  as a technical device  to apply gradient descent to the   Mumford-Shah functional~\cite{Mumford89}. Over the years, it has been extended in  numerous ways to address a rich variety of visual  applications. Earlier works include Shah and colleagues~\cite{Shah91,Shah96,ShahPien96,Pien97}, March and Dozio~\cite{March97}, Proesman, Pauwels and van Gool~\cite{Pauwels94}, Teboul {\sl et al.} ~\cite{Teboul98}. 
During the last couple of years we have witnessed an increasing number of promising works modifying or extending Ambrosio-Tortorelli/Mumford-Shah based models. Some examples are: Bar, Sochen and Kiryati~\cite{Bar06},  Rumpf and colleagues~\cite{Droske05,Droske07},  Erdem, Sancar-Yilmaz and Tari~\cite{EST07}, Patz and Preusser~\cite{Patz10}, Jung and Vese~\cite{JungVese09}.  These works together with many others collaboratively established  the role of AT function in variational formulations that jointly involve region and boundary terms.  

In the majority of the works,  the AT function  serves as an auxiliary variable to facilitate discontinuity-preserving smoothing and boundary  detection. Relatedly, the interaction  parameter is chosen sufficiently small  to better  localize boundaries. 
In contrast, Shah and Tari, starting  with~\cite{Tari96,Tari97} in late 90's, have focused on the  ability of the AT function in coding morphologic  properties of  shapes, regions construed by  boundaries.  Relatedly, they have weakened boundary/not-boundary separation   either  by choosing a  large interaction parameter or by  other means~\cite{Aslan05} and  focused on the 
geometric properties of the level curves  after constructing  the AT function  (reviewed in~\cite{Shah_chapter})  for shapes as: 
\begin{eqnarray}
 \underset{v}{\operatorname{{arg\,min}}}  \iint\limits_\Omega   \frac{1}{ \rho} \underbrace{ {\left(v({\mathbf x}) - {\chi}_\Omega ({\mathbf x})\right)}^2}_{\begin{smallmatrix}
  \text{boundary/interior}\\
  \text{separation}
\end{smallmatrix}} + \rho  \underbrace{ |\nabla v({\mathbf x})|^2}_{\begin{smallmatrix}
  \text{local interaction}
\end{smallmatrix}}
    \, \mathrm{d}x\,\mathrm{d}y  \nonumber  \\
  {\mbox{ \hskip 0.3cm with }} v({\mathbf x}) =0  \mbox{ for } {\mathbf x}=(x,y)\in {\partial \Omega}   \label{eq:ATenergy}
\end{eqnarray}
where $\Omega \in {\mathbf R^2}$  is  a  bounded open set  with a boundary  $\partial\Omega$ (denoting a shape);  ${\chi}_\Omega ({\mathbf x})$ is the shape indicator function which attains $1$ in $\Omega$ and $0$ on $\partial \Omega$;   $\rho$ is the  parameter.  The  first term forces strong boundary/interior separation while the second one  forces smoothness. 

The AT function of shape is related to a variety of morphological concepts. For instance,   it is a  weighted distance transform~\cite{Kimmel96,MaragosButt2000} with its level curves approximating  curvature-dependent motion~\cite{Osher88,Kimia95}.  Thus, it enables extraction of local symmetries and skeletons directly from grayscale images; that is, it  bridges image segmentation and shape description.  The ability of  level curves  in coding morphological information is  also exploited  by Droske and Rumpf~\cite{Droske07} to measure equivalence of two shapes in a registration problem.  

In this paper, following Shah and Tari~\cite{Tari96,Tari97,Aslan05}, we explore and extend the ability of an AT-like field  in coding features that are at a higher level than boundaries.   {Whereas the previous works   focus on local symmetry axes,  we   focus on shape's intuitive components as coded via upper and lower level sets. Our constructions are based on a new field obtained as the minimizer of a modified AT energy.  We discuss the geometry of the level curves of the new minimizer and exploit it to   extract a  part hierarchy tree endowed with a  probabilistic structure. }

The considered modification involves an additive augmentation of the interaction  term with a non-local term in a way  that the upper and lower zero level sets of the minimizer yield disjoint domains~\cite{Tari09} within which the minimizer is morphologically equivalent to the AT function.   
Following the pioneering work of Buades, Coll and Morel~\cite{NLMeans}, UCLA group formulated interesting non-local variational formulations, including  non-local versions of the Ambrosio-Tortorelli/ Shah approximations of the Mumford-Shah functional~\cite{JungVese09,Jung09} by replacing  local image derivatives with non-local ones. This kind of modification is very different from our modification which modifies the phase field itself.     

In this paper, we  focus on shapes.  {Nevertheless, the long term goal of our work is to bridge low level processes such as segmentation and image registration with the high level process of shape abstraction. Integration of the presented developments to Mumford-Shah type models via coupled PDEs framework is a future work. }
\section{A Modified Energy and Its Minimizer}
\label{sec:newminimizer}
 Let us consider
\begin{eqnarray}
\iint\limits_\Omega
 \frac{1}{\rho}  {\left( \omega({\mathbf x})-f ({\mathbf x}) \right)}^2 
+ 
 \rho 
\left[  |\nabla \omega({\mathbf x})|^2 + \left( {\mathbf E_{\mathbf x \in \Omega} \omega(\mathbf x)} \right)^2 \right] 
  \, \mathrm{d}x\,\mathrm{d}y    \nonumber \\ 
\mbox{ with } \omega({\mathbf x})=0 \mbox{ for } {\mathbf x} = (x,y) \in
{\partial \Omega} \hskip 2cm \label{eq:newenergy}
\end{eqnarray}

where $ \mathbf E_{\mathbf x \in \Omega} \omega(\mathbf x)$  is the expectation of $\omega$ given by $\frac{1}{| \Omega | }  \iint \omega({\mathbf x}) \, \mathrm{d}x\,\mathrm{d}y $ and $f ({\mathbf x}) $  is the distance transform.  The new energy to be minimized is composed of three terms and  obtained by modifying the AT energy in (\ref{eq:ATenergy}) in two aspects. 

Firstly, the interaction term of  (\ref{eq:ATenergy})    is additively augmented with $\left( {\mathbf E_{\mathbf x \in \Omega} \omega(\mathbf x)} \right)^2$. This new term forces the minimizer to acquire a low average value with the average being  computed over the entire domain. 
At a first glance, this  seems to favor spatial homogeneity by forcing the minimizer to attain values close to zero.  Yet, the minimum of $ \iint \left( {\mathbf E_{\mathbf x \in \Omega} \omega(\mathbf x)} \right)^2 \mathrm{d}x\mathrm{d}y$ is also reached when $\omega$ oscillates, that is, when it attains both negative and positive values  adding up to $0$.  In this respect, the third term  is  a separation term partitioning $\Omega$ into subdomains  of opposing signs. Due to the influence of the   $ | \nabla . |^2$ term which penalizes spatial inhomogeneities, locations of  identical sign  tend  to form spatial groups.  
Obviously, the minimizer of $ \iint_\Omega{ \left[  |\nabla \omega({\mathbf x})|^2 + \left( {\mathbf E_{\mathbf x \in \Omega} \omega(\mathbf x)} \right)^2 \right] \mathrm{d}x\mathrm{d}y}$  subject to homogenous Dirichlet boundary condition is the flat function $\omega=0$  unless accompanied by an external inhomogeneity. 

Indeed, the  purpose of the second modification is to influence   spatial grouping of positive and negative values of $\omega$ in a particular way that the sign change separates the {\sl gross} structure from the boundary detail. In particular, the upper zero level set  $ \left\{ \Omega^{+}= (x,y) \in\Omega:  \omega(x,y) > 0  \right\} $  covers central regions  whereas the lower zero level set $ \left\{ \Omega_{-}= (x,y) \in\Omega:  \omega(x,y) < 0  \right\} $ covers peripheral regions  containing limbs, protrusions and boundary texture or noise. 
Towards this end,  the indicator  ${\chi}_\Omega ({\mathbf x})$  is replaced   by  a weighted indicator  that is  a monotonically increasing function of the shortest distance to the boundary, namely, the distance transform.  
 {As before, the first term favors separation of the domain into phases; however,  the phases are the level curves of the distance transform. }
 {Since, however, the level curves of the AT function in (\ref{eq:ATenergy}) are equivalent to the level curves of a smooth distance transform~\cite{Tari97},  this change  merely  scales $\omega$  without qualitatively affecting the geometry of its  level curves. } Nevertheless, when the   terms considered together, the minimizer tends  to have positive values at central locations and negative values at peripheral locations because the penalty incurred by assigning negative values to central locations with higher positive $f$ values  is higher than the penalty incurred by assigning negative values to locations with lower $f$ values.

Similar to  the AT function, the new minimizer  is  a compromise between  inhomogeneity and homogeneity though  the inhomogeneity is forced both externally  (by $f$)  and internally (by the third term); or  it  is the best approximation of  an external inhomogeneity  $f$ subject to internal constraints.

The parameter   $\rho$ should be chosen large enough so  that  the attachment to the external inhomogeneity  should not dominate over the tendency to interact.  Indeed, in the absence  of the third term, a good practice is to chose $\rho$ at least on the order of the maximum thickness  for the   diffusive   effect of   $ | \nabla . |^2 $ to influence  the entire shape~(\cite{Aslan05}; Fig.~1 in~\cite{Tari97}).  The same argument also holds here
since the effect of the third term is to partition $\Omega$ into subdomains within which $\omega$ is morphologically similar to the AT function. 
Additionally, notice that the expression responsible for sign change, $\iint \omega({\mathbf x}) \, \mathrm{d}x\,\mathrm{d}y $,  has  been already normalized by $\frac{1}{| \Omega | }$. 
As such, $\rho$ should be larger than  $\sqrt{ | \Omega | }$.

\begin{figure}[htb]
\centering
\begin{tabular}{ccc}
\includegraphics[angle=0,width=0.25\linewidth, trim= 0 0 20 0, clip]{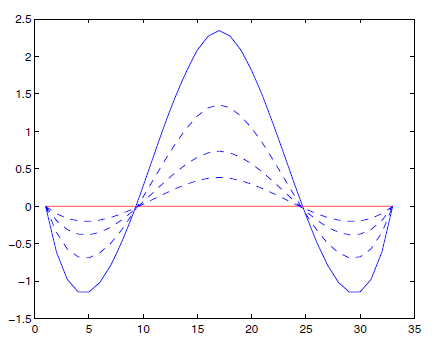}  &
\fbox{\includegraphics[angle=0,width=0.4\linewidth, trim= 80 80 50 80, clip]{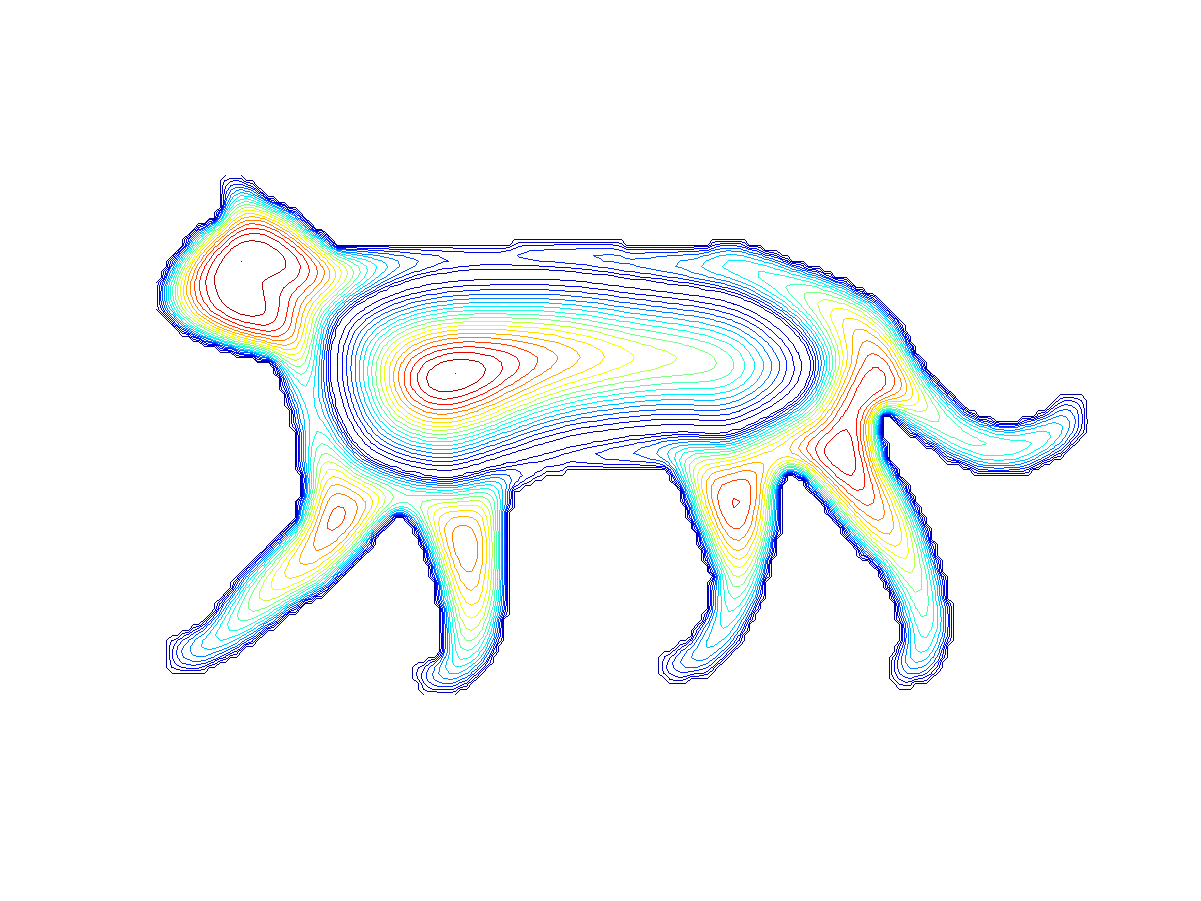} }&
\fbox{\includegraphics[angle=0,width=0.28\linewidth, trim= 130 50 100 30, clip]{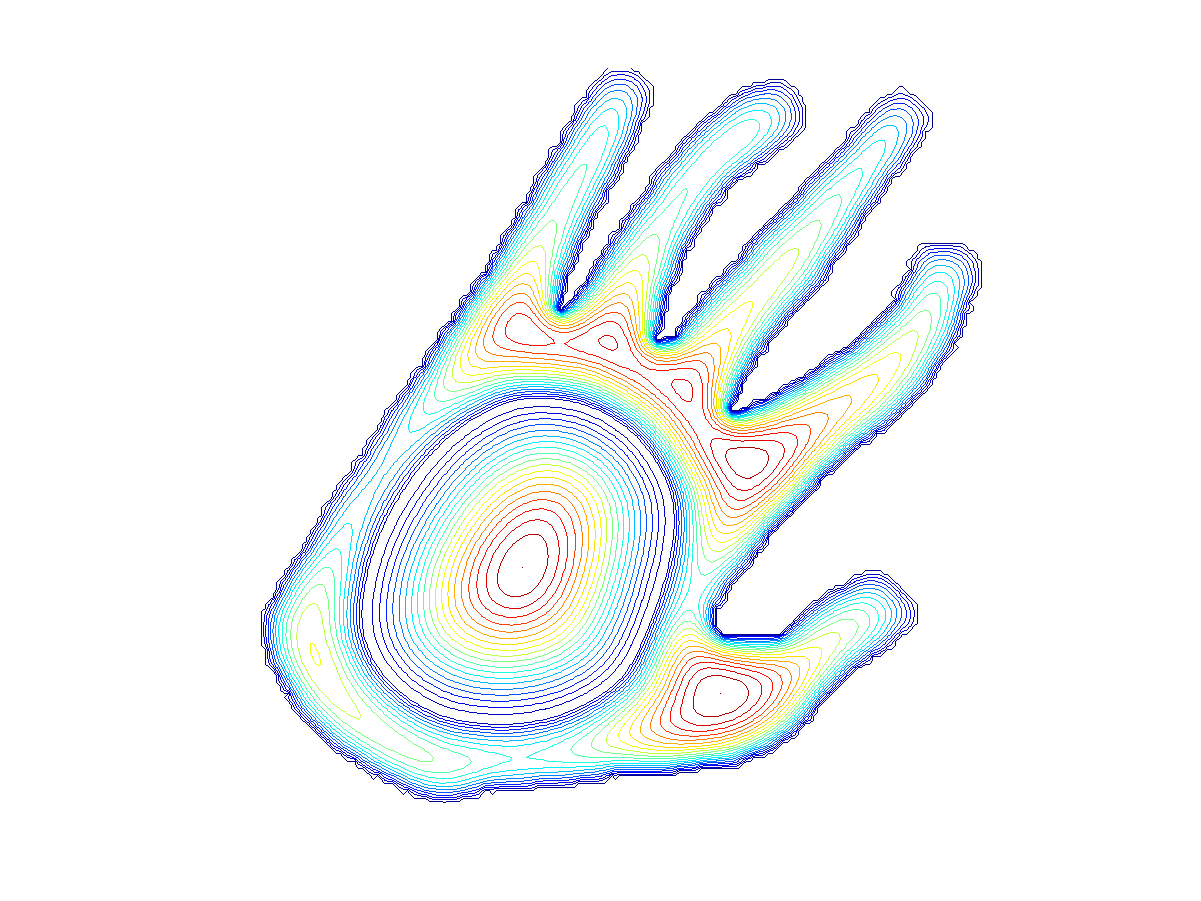} }  \\
(a) & (b) & (c) 
\end{tabular}
\caption {(a) $\omega$ for an interval for varying values of $\rho$ ranging between  $ \sqrt{ | \Omega | }$ and $ 0.5* | \Omega | $.  (b) Illustrative level curves of $\omega$. }
\label{fig:Fig1}
\end{figure}

In Fig.~\ref{fig:Fig1} (a),  an  illustration for a 1-D case is given:  $\omega$ is plotted for four different values of  $\rho$ ranging between  $ \sqrt{ | \Omega | }$ and $ 0.5* | \Omega | $.  
 Naturally,  $\omega$ gets flatter as $\rho$ increases.   (The flattening can be avoided by scaling either $f$ or $\omega$.) Nevertheless,  the locations of the extrema and the zero crossings remain the same unless $\rho$ is significantly smaller than $\sqrt{ | \Omega | }$.   
Similarly in 2-D,  the geometry of the level curves is stable as long as $\rho$ is chosen suitably large. 
 Illustrative level curves are depicted  in Fig.~\ref{fig:Fig1} (b-c). 
 Absolute values of $\omega$ separately normalized within regions of identical sign are used for convenience of color visualization.
Zero level curves separate  central and peripheral structures in the form of upper and lower zero level sets: $ \left\{ \Omega^{+}= (x,y) \in\Omega:  \omega(x,y) > 0  \right\} $ and 
$ \left\{ \Omega_{-}= (x,y) \in\Omega:  \omega(x,y) < 0  \right\} $. The peripheral structure includes all the detail: limbs, protrusions, and boundary texture or noise.   In contrast, the central structure is a very coarse blob-like form; it  can even be thought as an interval estimate of the center whereas the centroid is the  point estimate.

 Most commonly,  $\Omega^+$ is a simply connected set. Of course, it may also be either  disconnected or multiply connected.  For  instance, it is disconnected for a dumbbell-like shape  (two blobs of comparable radii combined through a  thin neck)  whereas it is multiply connected for an annulus formed by two concentric circles. Indeed, the annulus  gets split into three concentric rings where the middle  ring is the $\Omega^+$. 
{For quite a many shapes, however, $\Omega^+$ is a simply   connected set.  

Firstly, shapes  obtained by protruding a blob as well as  shapes whose peripheral parts are smaller or thinner than their main parts  always have  a  simply connected $\Omega^+$.  This is expected: When the width of a part is small, the highest value of  $f$ inside the part is small. That is, the local contribution to  $ {\left( \omega-f  \right)}^2 $ incurring  due to  negative values is less significant for such a part as compared to locations with   higher positive values of $f$.  Consequently,  $\omega$ tends to attain  negative values on narrow or small  parts as well as on protrusions.} Shapes  with holes also  have  a simply connected $\Omega^+$ as long as  the holes are far from the center. 

Secondly, even a dumbbell-like shape may have  a simply connected $\Omega^+$.      This happens if the join area, namely, the neck  is wide  enough. 
Nevertheless, this does not cause any  representational instability: 
Whereas the  $\Omega^+$ for a blob-like shape has a unique maximum located roughly at its centroid, the $\Omega^+$ for a dumbbell-like shape 
 has  two local maxima   indicating  two bodies. 
Each body is  captured by a connected component of an upper level set whose bounding curve  passes through a  saddle point.  
At a saddle point $\vec{p}, {\text{such that }} \omega(\vec{p})=s$,  the $s-level$ curve has a double point singularity, {\sl i.e.}  it forms a cross.   As such, the upper level set $ \left\{
\Omega^{s}= (x,y) \in\Omega^+:  \omega(x,y)> s  \right\} $ 
yields two disjoint  connected components capturing the two parts of  the central structure.

In contrast to $\Omega^+$, the peripheral structure $\Omega_-$ is often  multiply connected. Indeed, its  hole(s) are carved  by $\Omega^+$. 
It is also  possible  that $\Omega_-$  is disconnected.   
For instance,  for an annulus, it  is two concentric rings.  Additionally, $\Omega_-$  may be disconnected   when there are  several elongated limbs organized around a  rather small central body,  e.g.,  a palm tree.   $\Omega^+$, being small, is tolerated to grow and   reach to the most concave parts of the shape boundary creating a split of $\Omega_-$ by the zero-level curve. 
Similar to those in $\Omega^+$,  the level curves in $\Omega_-$ that are passing through saddle points  provide further   partitioning.  
The partitions are in the form of  lower level sets 
  $ \left\{
\Omega_{s}= (x,y) \in\Omega_-:  \omega(x,y)
< s  \right\} $.    

To sum up, within both $\Omega^+$ and $\Omega_-$,  nested open  sets (upper level sets inside  $\Omega^+$ and lower level sets inside $\Omega_-$)  characterize the domain.  
The level curves bounding the level sets are either closed curves or closed curves with crossing points.  The ones  with crossing points are of particular interest because  the respective level set  is partitioned at those points  into two  distinct connected components. A crossing of a level curve occurs at a saddle point of $\omega$. 
Of course, each lower level set may contain other saddle points. Consequently, the partitioning is binary and iterative and determined by the order of saddle points.  

It is not generically possible that a level curve has singular points of higher order  because such singular  points  are unstable and may be removed by a slight change in $\omega$.    It is also highly unlikely that a connected component of an $s-level$ curve has two distinct crossing points. 
 This issue is tackled in \S\ref{sec:random} via randomization.    
\section{ Randomized Hierarchy Tree}
\label{sec:random}
\begin{figure}[htb]
\centering
\begin{tabular}{c}
{\includegraphics[angle=0,width=0.8\linewidth,  trim= 0 650 0 50, clip]{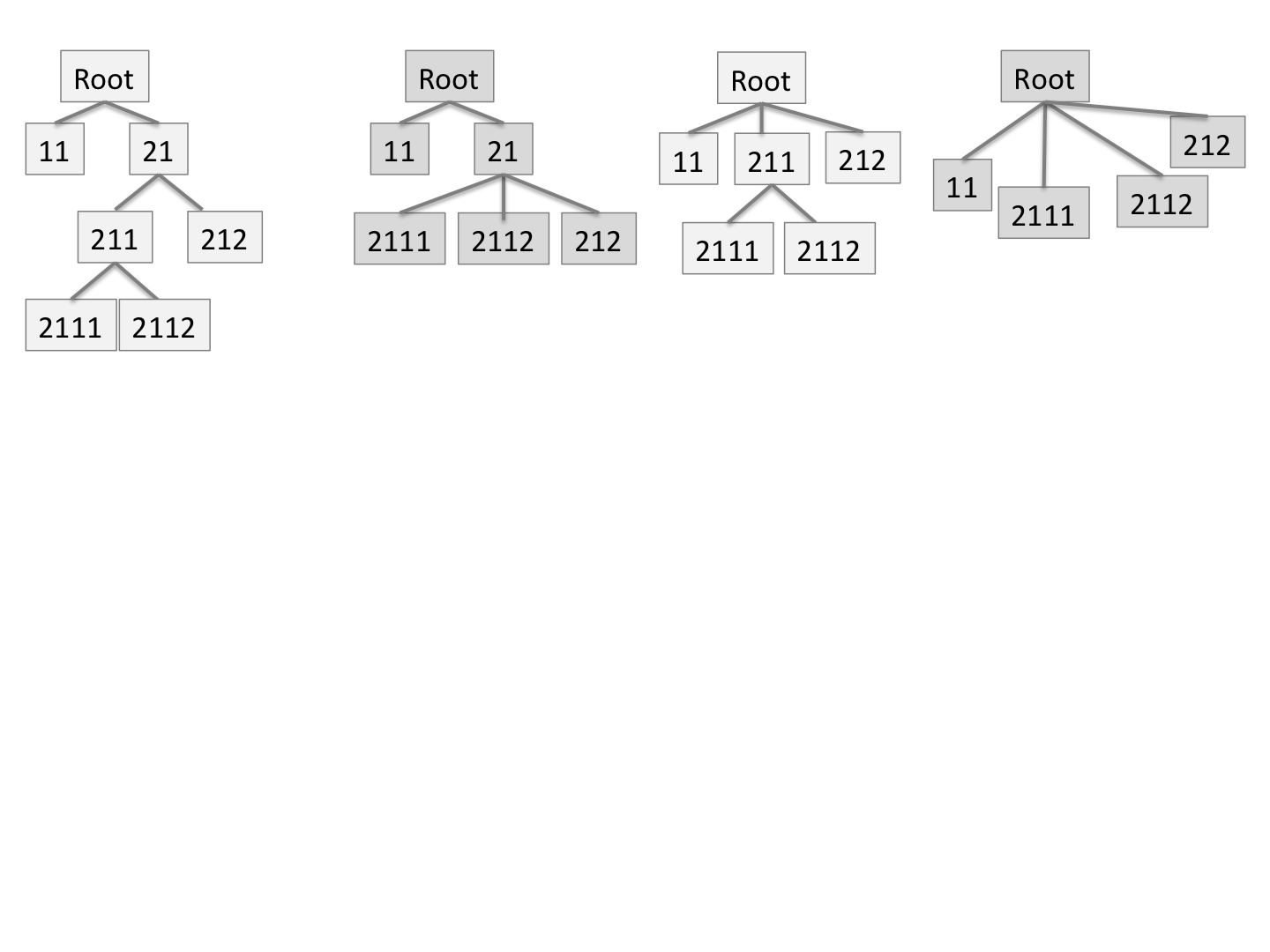}} \\
\mbox{  } \hskip 0.8cm (a) \mbox{  } \hskip 2cm  (b) \mbox{  } \hskip 1.6cm  (c)  \mbox{  } \hskip 1.8cm   (d) \mbox{  } \hskip 1cm \mbox{ } 
\end{tabular}
\caption { (a) An initial part  tree (a) and  its possible re-organizations (b-d). A random sample should be in one of the four forms. See the text.}
\label{fig:tree_demo}
\end{figure}

Since  the partitioning  inside both  the $\Omega^+$ and   $\Omega_-$ of a shape are iterative and binary, the parts can be organized starting from the second level  in the form of a proper binary  tree. Let the shape be the root node and its  children be  the upper and lower zero level sets, namely, the disjoint regions of  the central and  peripheral structures. 
Suppose  the  central and peripheral structures  are   respectively composed of $N_c$ and $N_p$ disjoint sets.    Let us  enumerate the nodes   holding these sets  as $11,12,\cdots, 1N_c$ for the $\Omega^+$ and  as   $21, 22,\cdots, 2N_p$ for the $\Omega_-$.  This is the second  level of the tree and the first level of the partitioning. 
Of course, the root may have more than two children. Nevertheless,  starting from the children of the root, 
each subtree is a  proper binary tree because all the  splits inside an $\Omega^+$  or  $\Omega_-$  occur at saddle points; that is, 
each connected component of the second  level and its children either get  split into two level sets or remain as they are.   We call this hierarchical organization  as the  {\bf Initial Part Tree}.  
A  hypothetical   initial part tree is illustrated in  Fig.~\ref{fig:tree_demo} (a).  In a real example, the nodes  hold application dependently selected  attributes of the respective level sets. 

Binary splits according to saddle points produce  collections of parts which are at the leaf level   consistent across visual changes. However,  the hierarchical order  and granularity of parts are not necessarily consistent.
For instance,  a weak saddle is easily removed when the shape is slightly smoothed.  Likewise, certain non-generic configurations  such as level curves with spatially distinct saddle point singularities or triple point singularities cannot occur; indeed, such configurations  are easily replaced by one of the corresponding generic configurations which may differ for similar  shapes.  
 Furthermore,  when a shape is occluded by another shape, added peripheral parts change the positions  of some of the previous level sets in the hierarchy.  
Nevertheless, the relative values of $\omega$  at saddle points prompting consecutive splits are stable indicators of the organizational hierarchy.    Of course, attempting to convert a saddle point value  to a tree depth by discretization  brings back the previous robustness issue. 

Instead, we  use the difference between the values of two successive  saddle points as a measure of saliency  for the partitioning prompted by the latter saddle point.  Converting the saliency measure to a probability measure and considering probability measures for all nodes,  we endow the initial part tree with a random  structure from which possible re-organizations of the initial hierarchy tree are to be sampled.  We call the new structure as the {\bf Randomized Part Hierarchy Tree}. Below, we give the details of the randomization procedure. In contrast to the respective  initial part tree, a random sample from a randomized part hierarchy tree is not necessarily a proper binary tree.

The randomization  starts from  level $3$ nodes and propagates through their children.  Recall that this is the first level of nodes that are created  via saddle points. For each pair of siblings,  there are two possible events:  
 The pair of siblings  either maintain their depth (no change in the local tree structure) or  inherit the depth of their  parent (change in the local tree structure). In the latter case, the node and  its sibling  replace their parent and become the children of their grandparent.  
The probabilities of the two events are derived from a quantity which we denote by $D\omega$. It is a property of a split meaning that the $D\omega$ values of  a pair of siblings are identical. Specifically,  it is  the  difference between the saddle point values of a node  and  its  parent divided by the saddle point value of the node.
Because the magnitude of the saddle point value  of  a node is always greater than that of  its parent, $0 < D\omega \leq 1$; the  equality is  attained at level $3$. 

 A small value of   $D\omega$ implies that the consecutive saddle points are closer in value; that is,  a  slight change in their value changes their  order hence the local tree structure.  Equivalently a  large value of   $D\omega$ implies that the consecutive saddle points are well separated,  therefore,  the local structure  is stable. We require that the probability $p$ that a local structure change is necessary  approaches  $1$ as  $D\omega$ approaches  to the smallest possible value which is $0$.  Equivalently,   $p$ should approach  $0$ as  $D\omega$ approaches to its largest possible value. 
The  function $e^{-4D\omega}$ is a good  candidate for estimating $p$; there is less than $2\%$ chance for reorganization since $e^{-4}=0.018$ for the largest  possible   $D\omega$.

 Let us consider the initial part tree in Fig.~\ref{fig:tree_demo} (a). Assume that $D\omega=0.301$ for  nodes $211$ and $212$. 
With probability $(1-p)=0.7$,    the local structure is preserved,   whereas with probability $p=0.3$  nodes $211$ and $212$  replace their parent and become children of their grandparent, the  root. 
Assume that $D\omega=0.128$ for  $2111$ and $2112$. Then with probability $(1-q)=0.4$,    the local structure is preserved,  while with probability $q=0.6$ nodes   $2111$ and $2112$  replace their parent and become children of their grandparent  which is either node $21$ with  $(1-p)=0.7$  or the root with $p=0.3$.  
Thus, there are four possible organizations: 
With probability $(1-p)(1-q)=0.28$, the entire structure is preserved. 
With probability $(1-p)q=0.42$, the tree is re-organized as in (b).
With probability $p(1-q)=0.12$, the tree is re-organized as in (c). 
With probability $pq=0.18$,  the tree is re-organized as in (d). 
\section{Experimental Results and Discussion}
To evaluate the  effectiveness of endowing the part  hierarchy tree with a probabilistic structure,   we consider  a pairwise matching  problem. It is formulated as finding a maximal clique in the joint association graph of the pair of trees  to be matched, e.g. ~\cite{Pelillo}.    At each   experiment, each of the two randomized part hierarchy trees is  independently  sampled several times and then all the sample pairs are matched.  The trees in  the sample pair  that yields  the highest matching score are called as the winning  re-organizations. 

Depending on the application, various properties related to      the level sets stored at  nodes can be used as node attributes.    For instance, we extract   parts   enclosing each of the stored level sets  and then  use   their  area and  the maximum  $\omega$ values inside them  as  attributes.  We remark that the maximum value of $\omega$  is related to the part  width for the finest parts. 
Because  boundaries of  stored  level sets pass through  saddle points, an  enclosing part is  easily obtained  as a morphologic watershed zone,  whose seed is the respective level set~\cite{Meyer94}. 
Both to keep  illustrations simple and resource  requirements  low, we require that each part   to neighbor  the central structure and each seed and part to  have a certain size. Specifically,  splits are performed only through the saddle points that reside on the boundaries of the  watershed zones that are touching to the closure of the  central structure and any split  producing  a  seed  that is less  than $0.05\%$ of the shape  or a part that  is less  than $0.5\%$ of the shape  is ignored. 

We present four illustrative examples.   In each case,  correct associations are found despite   several order and granularity inconsistencies  resulting from  occluders, non-generic splits and weak saddles. 

\begin{figure}[h!tb]
\centering
\begin{tabular}{c}
\fbox{\includegraphics[angle=0,width=0.9\linewidth,  trim= 0 0 0 0, clip]{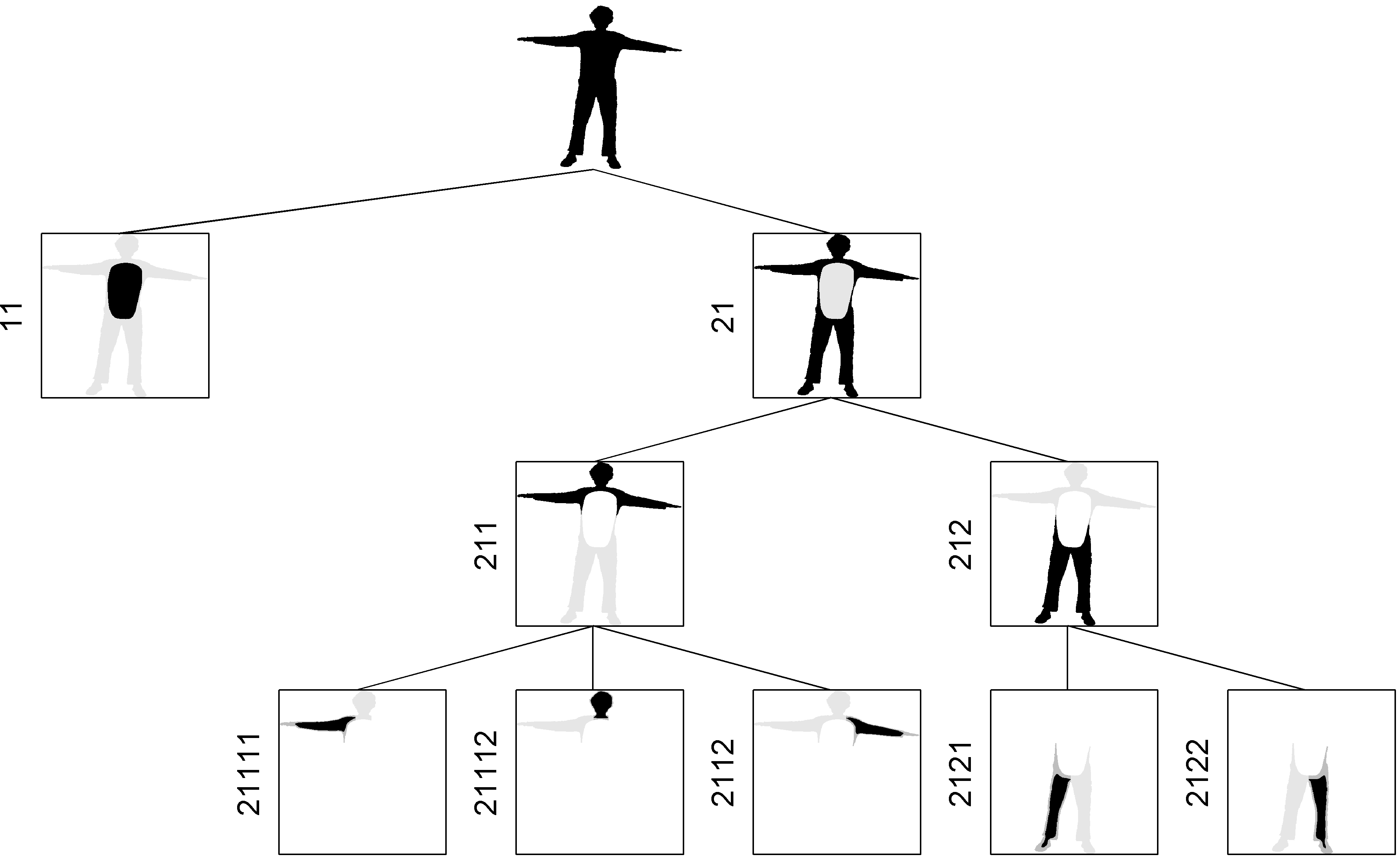}} \\
\mbox{} \\
\fbox{\includegraphics[angle=0,width=0.9\linewidth,  trim= 0 0 0 0, clip]{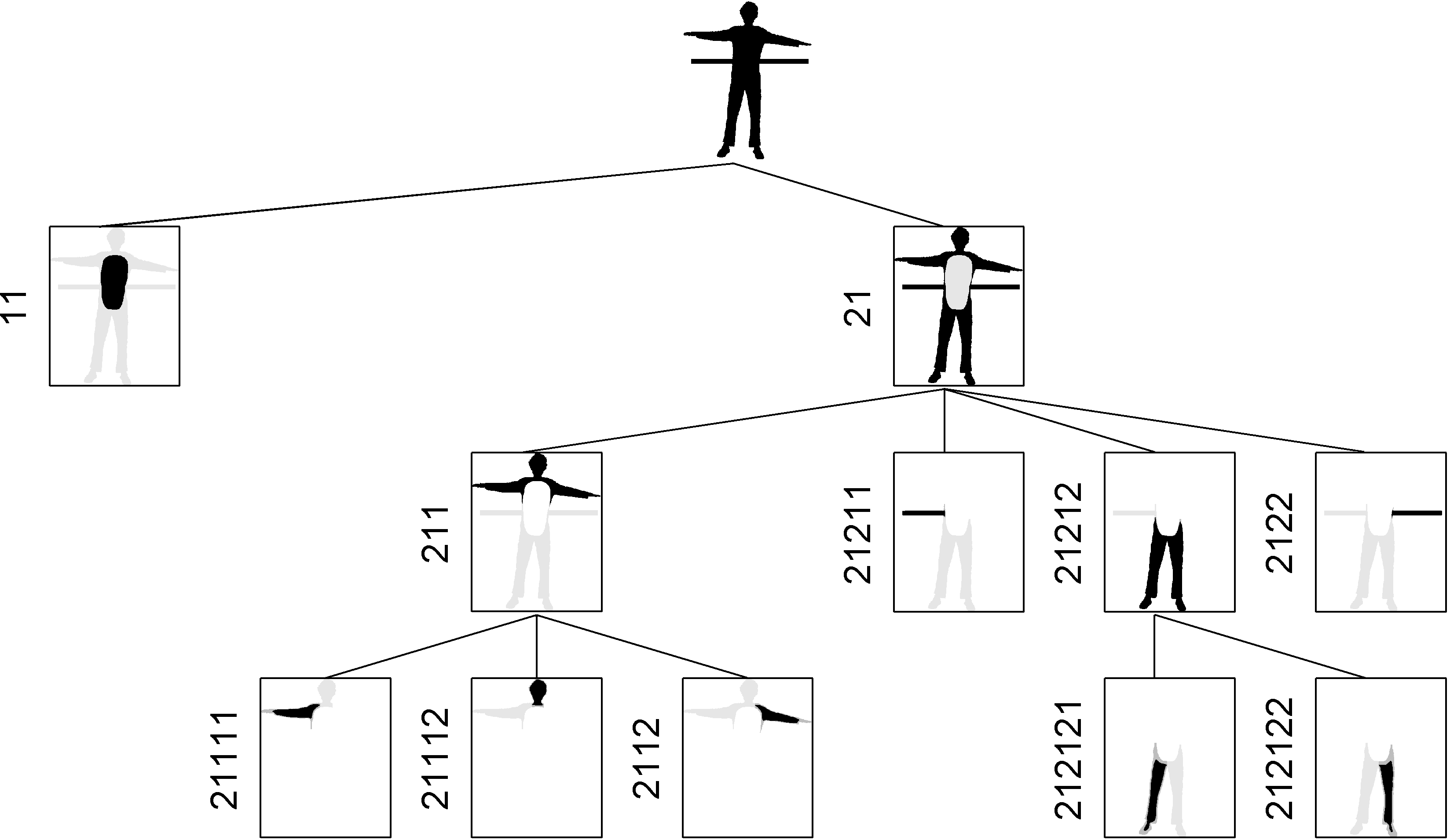}} \\
\end{tabular}
\caption {The winning re-organizations for two shapes.   The numbering of the nodes reveal their order in the initial binary tree. See the  text. } 
\label{fig:sampled_kadin}
\end{figure}

The first example is a matching between a human silhouette and its occluded version. The winning re-organizations  for each of the two  trees  are  shown  in Fig.~\ref{fig:sampled_kadin}.   At each node, the watershed (enclosing part) is depicted as  dark gray and the respective  level set (seed part)  depicted as   black is  superimposed on the part; the neighboring part (the light gray) is  also shown  even though it is not used for the matching.
Due to page limits, we cannot  provide the initial trees and probability distributions, but   the numbering of the nodes already reveals the structure of the initial binary tree. 

Firstly, notice  that the arms on the left and on the right  reside at different levels in both of the initial trees  as revealed by their  respective five versus four digit node numbers.
Ideally,  the almost symmetric upper bodies (nodes $211$)  should contain two distinct saddle points  $\vec{p_1}$ and $\vec{p_2}$ such that $\omega(\vec{p_1})= \omega(\vec{p_2}) =s$; that is, two distinct saddle points on a  single $s-level$ curve should  simultaneously yield  the three nodes: $21111, 2112$ (arms) and  $21112$ (head).   However, certain configurations  including this one  are  not generic; even the slightest perturbation imposes a strict order on the saddle points. Thus, firstly,  the combination of the head and either one of the arms is separated from the other arm  then the head-arm combination is partitioned.  Nevertheless, in each case, the saddle point value separating the head and arm on the left combination  from the arm on the right is   very close to the saddle point value  separating the head from the arm on the left; e.g.,  for the first shape, the respective saddle point values after normalization with respect to the global maximum of $\omega$ are  $-0.683$ and $-0.687$ while the saddle point value separating the entire  upper body from the entire lower body is $-0.053$.  Clearly,  the hierarchical order between the upper body and its children is more stronger than the hierarchical order among its children. 
We remark that  even though the  arm  re-organization is  not necessary for finding  correct part correspondences  since the structures of the upper bodies are  already  the same for the two  initial trees, the left and right arms are brought to the same level.  This is  because the  probabilities of retaining the  initial binary  local structures is very low due to the closeness of the consecutive saddle point values.

Secondly, notice that the legs of the occluded figure are  at the sixth level whereas the legs of the un-occluded one are at the fourth level, as revealed by their node numbers.  This is due to the influence of  two additional parts (watershed regions)  belonging to  the occluder  and poses a challenge for the tree  matching.   Nevertheless,  the legs are brought to  the same level as well as  all of the corresponding parts 
and  correct  associations are found:
$11 \Leftrightarrow 11$ (central regions),
$211 \Leftrightarrow 211$ (upper bodies), 
$21111 \Leftrightarrow 21111$ (arms on the left), 
$21112 \Leftrightarrow 21112$  (heads), 
$2112 \Leftrightarrow 2112$  (arms on the right), 
$212 \Leftrightarrow 21212$  (lower bodies)
$2121 \Leftrightarrow 212121$  (legs on the left)
$2122 \Leftrightarrow 212122$  (legs on the right).    

\begin{figure}[htb]
\centering
\begin{tabular}{cc}
\fbox{ \includegraphics[angle=0,width=0.5\linewidth,  trim= 25 20 20 10, clip]{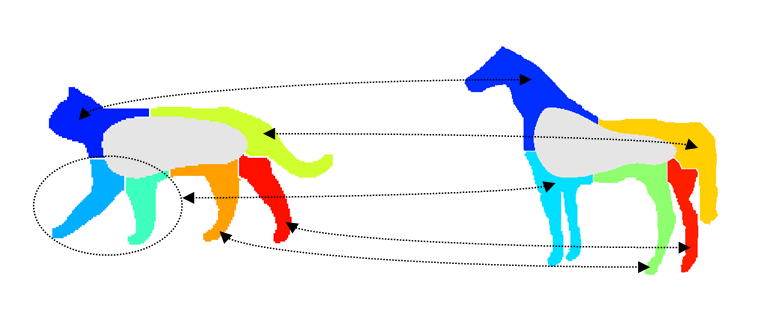} } &
\fbox{ \includegraphics[angle=0,width=0.35\linewidth,  trim= 75 60 60 80, clip]{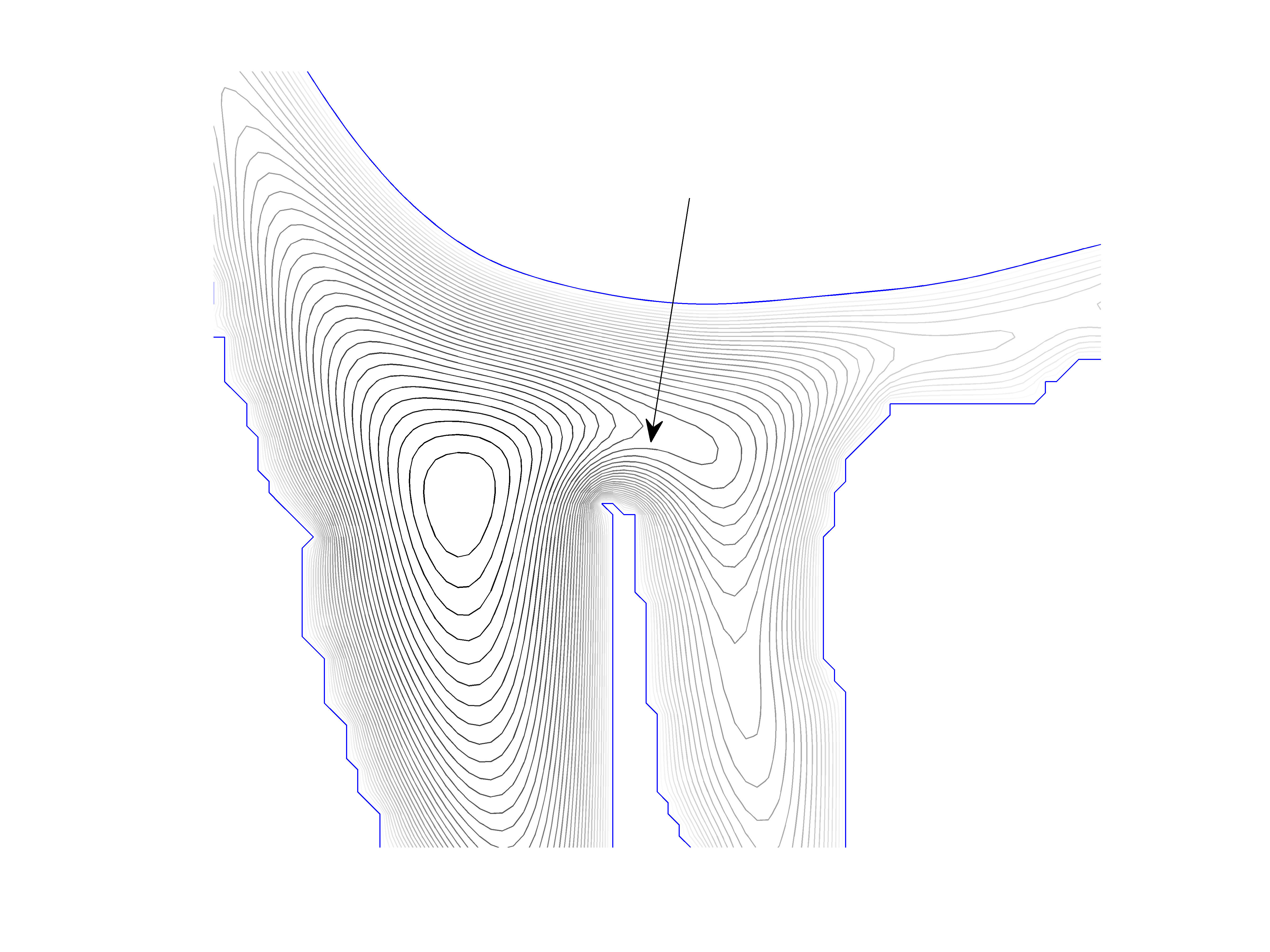}}
\end{tabular}
\caption{A granularity inconsistency. Due to the weak saddle marked by the arrow in the left,  the part of the horse corresponding to its front legs  cannot be further partitioned. Nevertheless, it is correctly associated to a non-leaf node of  the cat.}
\label{fig:matching1}
\end{figure}

 In  the next three examples, due to limited space, only the matchings  where at least one member of the matching pair is a leaf are depicted even though  entire hierarchical structures are matched. Non-leaf nodes are  circled.  These examples also demonstrate the necessity of not restricting the correspondence search to leaf nodes.  The matching between a cat and a horse in Fig.~\ref{fig:matching1}  illustrates a granularity inconsistency. Due to a weak saddle marked  by the arrow in the left, the front legs of the horse are not further partitioned. Nevertheless, this   inconsistency is resolved by matching the respective leaf node of the horse tree to a  non-leaf node of the cat tree, the parent of the  two nodes each holding a front leg of the cat. 

\begin{figure}[htb]
\centering
\begin{tabular}{cc}
\fbox{ \includegraphics[angle=0,width=0.46\linewidth,  trim= 25 20 20 10, clip]{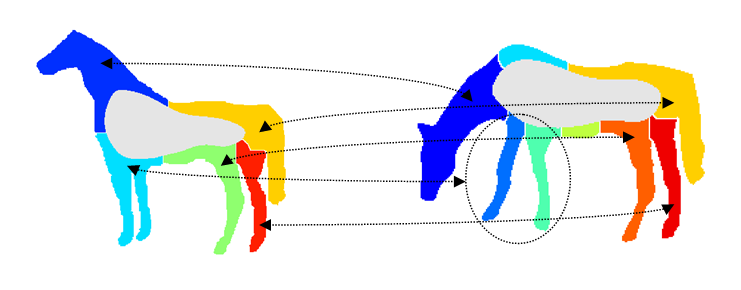}} & 
\fbox{\includegraphics[angle=0,width=0.46\linewidth,  trim= 20 10 20 10, clip]{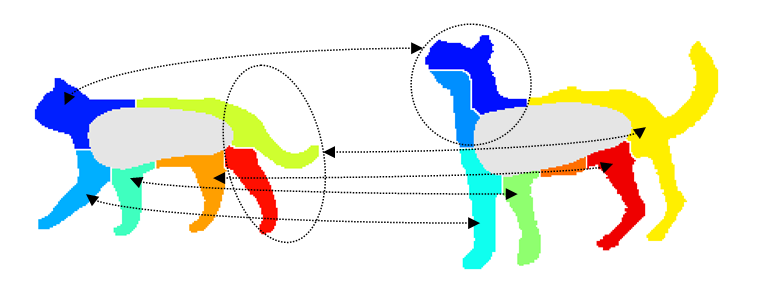}} \\
(a) & (b)
\end{tabular}
\caption {Two more cases involving  both level and granularity inconsistencies.}
\label{fig:matching2}
\end{figure}

Fig.~\ref{fig:matching2} (a) depicts the matching of the same horse to another horse. In addition to the previous granularity inconsistency,   there are  several order inconsistencies which are not noticeable at the leaf level presentation.  For instance, the rear body of the first horse firstly  splits into the fourth leg and tail versus the third leg,  and then the fourth leg is separated from the tail. On the other hand, the rear body of the second horse after a spurious division gets   splits into the rear  legs versus the tail, and then the two legs are separated. Consequently, a two level difference between the third leg of the first horse (node $2121$) and the third leg of the second horse (node $212221$) is formed. Despite both granularity and order  inconsistencies, all of the parts are correctly matched.  

The final example    (Fig.~\ref{fig:matching2} (b))  involves several difficulties due to three  weak saddles  resulting with three unintuitive partitions for the second cat: Firstly,   its head  is fragmented; secondly, its rear body goes through a spurious division causing  an erroneous shift in the levels of  its sub-parts; thirdly,   its  fourth leg and  tail are  not separated.  Nevertheless, the selected clique contains all of the correct associations.  
The rear body and its parts for the second cat are properly lifted one level up; consequently, the correct associations of the parts of the rear bodies are  found successfully.  The head of the first cat  matches to the parent of the two leaves holding two unintuitive parts  of the head of the second cat.   The two  head fragments of the second  cat as well as the    fourth leg and the tail  of the first cat  are correctly excluded from the selected  clique as there are no corresponding parts in the other tree.
%

%
%

\subsection*{Acknowledgements} ST is supported by the   Alexander von Humboldt Foundation.  She  extends  her gratitude to  the members of Sci. Comp. Dept. of Tech. Universit\"at M\"unchen, in particular to Folkmar Bornemann,  for providing a wonderful  sabbatical stay in every respect.  MG  is funded as   MS student via T\"ubitak project 108E015 to ST. 

\bibliographystyle{plain}
\bibliography{ssvm11}

\end{document}